\documentclass[final]{cvpr}

\usepackage{times}
\usepackage{epsfig}
\usepackage{graphicx}
\usepackage{amsmath}
\usepackage{amssymb}

\usepackage{caption}

\usepackage{graphics}
\usepackage[usenames,dvipsnames,table]{xcolor}
\usepackage{array}
\usepackage{multirow}
\usepackage{wrapfig}
\usepackage{hhline}
\usepackage{pifont}
\newcommand{\showimagew}[2][\linewidth]{\includegraphics[width={#1}]{{#2}}}

\newcommand{\topscore}[1]{\textcolor{blue}{\textbf{#1}}} 
\newcolumntype{K}[1]{>{\centering\arraybackslash}p{#1}}
\newcommand{\negvspace}{\vspace{-0.2cm}}

\usepackage{enumitem}

\usepackage[pagebackref=true,breaklinks=true,colorlinks,bookmarks=false]{hyperref}

\def\cvprPaperID{9996} 
\def\confYear{CVPR 2021}
\newcommand{\yagiz}[1]{\textcolor{blue}{{[Ya\u{g}{\i}z: #1]}}}

\newcommand\blfootnote[1]{%
  \begingroup
  \renewcommand\thefootnote{}\footnote{#1}%
  \addtocounter{footnote}{-1}%
  \endgroup
}

\begin{document}

\title{Boosting Monocular Depth Estimation Models to High-Resolution via Content-Adaptive Multi-Resolution Merging\negvspace\negvspace}


\author{
S. Mahdi H. Miangoleh$^{*1}$\quad 
Sebastian Dille$^{*1}$\quad
Long Mai$^{2}$\quad
Sylvain Paris$^{2}$\quad
Ya\u{g}{\i}z Aksoy$^{1}$
\negvspace\\\\\negvspace
$^{1}$ Simon Fraser University \quad\quad $^{2}$ Adobe Research
}


\twocolumn[{%
\renewcommand\twocolumn[1][]{#1}%
\maketitle
\begin{center}
    \centering
    \negvspace
    \negvspace
    \negvspace
    \showimagew[\linewidth]{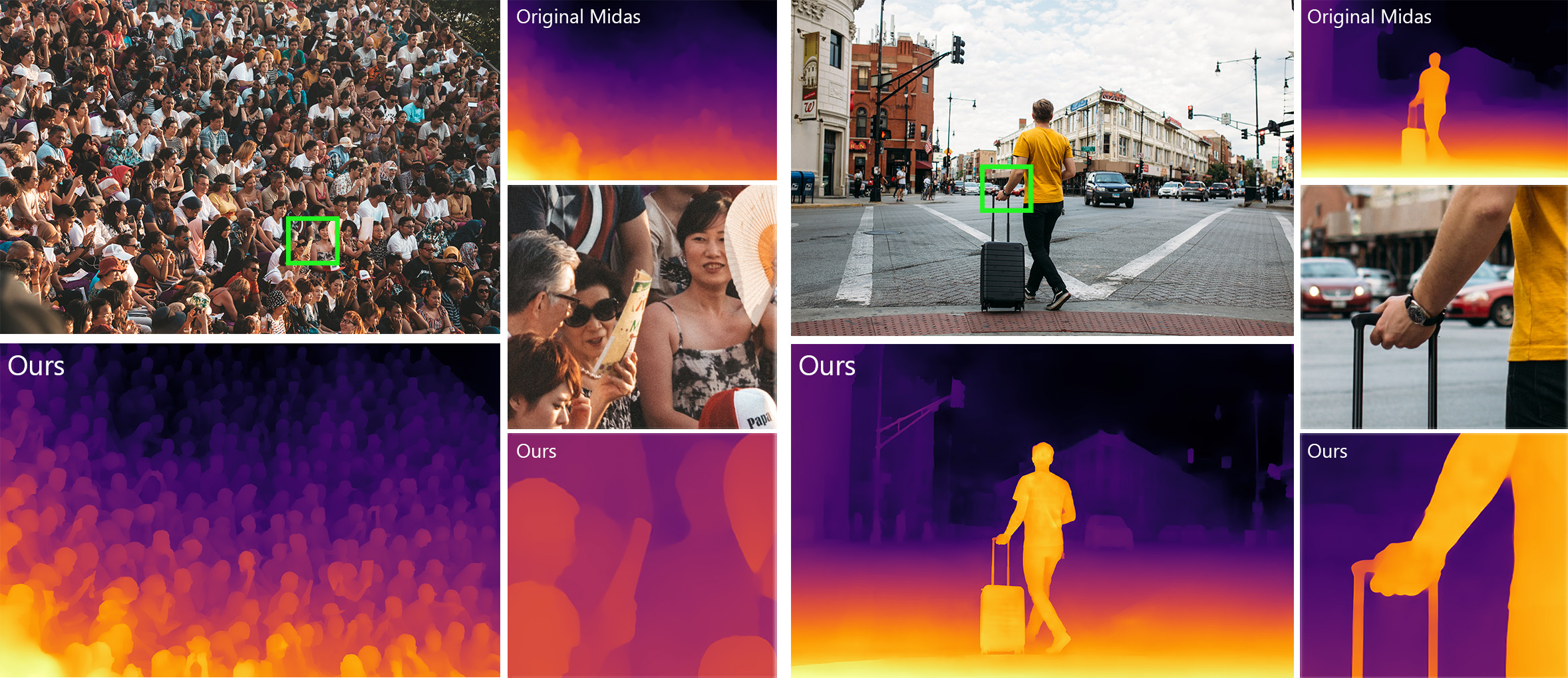}
    \negvspace
    \negvspace
    \negvspace
    \captionof{figure}{
      We propose a method that can generate highly detailed high-resolution depth estimations from a single image. Our method is based on optimizing the performance of a pre-trained network by merging estimations in different resolutions and different patches to generate a high-resolution estimate. We show our results above using MiDaS~\cite{midas} in our pipeline.
    \negvspace
    }
    \label{fig:teaser}
\end{center}%
}]

\begin{abstract}
\negvspace
Neural networks have shown great abilities in estimating depth from a single image. 
However, the inferred depth maps are well below one-megapixel resolution and often lack fine-grained details, which limits their practicality. 
Our method builds on our analysis on how the input resolution and the scene structure affects depth estimation performance. 
We demonstrate that there is a trade-off between a consistent scene structure and the high-frequency details, and merge low- and high-resolution estimations to take advantage of this duality using a simple depth merging network. 
We present a double estimation method that improves the whole-image depth estimation and a patch selection method that adds local details to the final result. 
We demonstrate that by merging estimations at different resolutions with changing context, we can generate multi-megapixel depth maps with a high level of detail using a pre-trained model.

\negvspace\negvspace
   \blfootnote{($^{*}$) denotes equal contribution.}
\end{abstract}

\section{Introduction}
\label{sec:intro}

\begin{figure*}\vspace{-0.1in}
\showimagew[\linewidth]{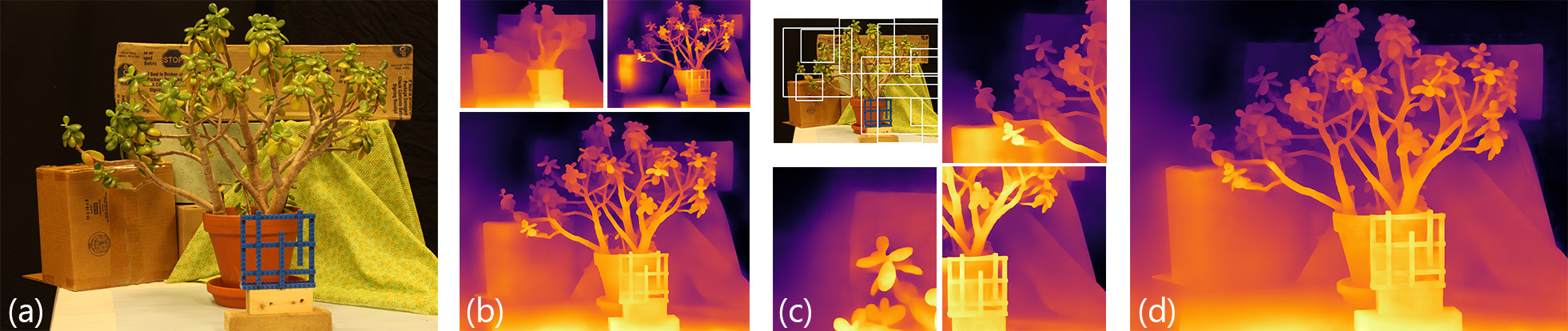}
\caption{
The pipeline of our method: (b) We first start with feeding the image in low- and high-resolution to the network, here shown results with MiDaS~\cite{midas}, and merge them to get a base estimate with a consistent structure with good boundary localization. (c) We then determine different patches in the image. We show a subset of selected patches with their depth estimates. (d) We merge the patch estimates onto our base estimate from (b) to get our final high-resolution result.
\negvspace
}\vspace{-0.12in}
\label{fig:pipeline}
\end{figure*}

Monocular or single-image depth estimation aims to extract the structure of the scene from a single image. 
Unlike in settings where raw depth information is available from depth sensors or multi-view data with geometric constraints, monocular depth estimation has to rely on high-level monocular depth cues such as occlusion boundaries and perspective. 
Data-driven techniques based on deep neural networks have thus become the standard solutions in modern monocular depth estimation methods \cite{eigen2014depth,fu2018deep,godard2017unsupervised,godard2019monodepth2,li2018megadepth}. 
Despite recent developments in the field including in network design~\cite{fang2020towards,huynh2020guiding,johnston2020self,Poggi_2020_CVPR}, incorporation of high-level constraints~\cite{semanticconquerdepth,yang2020d3vo,zhang2019pattern,zhu2020edge}, and supervision strategies~\cite{godard2017unsupervised,godard2019monodepth2,guo2018learning,jiang2018self,klingner2020self}, achieving high-resolution depth estimates with good boundary accuracy and a consistent scene structure remains a challenge. 
State-of-the-art methods are based on fully-convolutional architectures which in principle can handle inputs of arbitrary sizes.
However, practical constraints such as available GPU memory, lack of diverse high-resolution datasets, and the receptive field size of CNN's limit the potential of current methods.

We present a method that utilizes a pre-trained monocular depth estimation model to achieve high-resolution results with high boundary accuracy.
Our main insight comes from the observation that the output characteristics of monocular depth estimation networks change with the resolution of the input image. 
In low resolutions close to the training resolution, the estimations have a consistent structure while lacking high-frequency details. 
When the same image is fed to the network in higher resolutions, the high-frequency details are captured much better while the structural consistency of the estimate gradually degrades. 
We claim following our analysis in Section~\ref{sec:monocular} that this duality stems from the limits in the capacity and the receptive field size of a given model.
We propose a double-estimation framework that merges two depth estimations for the same image at different resolutions adaptive to the image content to generate a result with high-frequency details while maintaining the structural consistency. 

Our second observation is on the relationship between the output characteristics and the amount and distribution of high-level depth cues in the input. 
We demonstrate that the models start generating structurally inconsistent results when the depth cues are further apart than the receptive field size. 
This means that the right resolution to input the image to the network changes locally from region to region. 
We make use of this observation by selecting patches from the input image and feeding them to the model in resolutions adaptive to the local depth cue density. 
We merge these estimates onto a structurally consistent base estimate to achieve a highly detailed high-resolution depth estimation. 

By exploiting the characteristics of monocular depth estimation models, we achieve results that exceed the state-of-the-art in terms of resolution and boundary accuracy without retraining the original networks. 
We present our results and analysis using two state-of-the-art monocular depth estimation methods~\cite{midas,xian2020sgr}. 
Our double-estimation framework alone improves the performance considerably without too much computational overhead while our full pipeline shown in Figure~\ref{fig:pipeline} can generate highly detailed results even for very complex scenes as Figure~\ref{fig:teaser} demonstrates.

\section{Related Work}
\label{sec:related}

Early works on monocular depth estimation rely on hand-crafted features designed to encode pictorial depth cues such as object size, texture density, or linear perspective~\cite{Saxena2008}. 
Recent works leverage deep neural networks to learn depth-related priors directly from training data \cite{chen2016single, eigen2014depth, godard2017unsupervised, Ramamonjisoa_2020_CVPR, wang2020cliffnet, wong2019bilateral, zheng2018t2net}. 
In recent years, impressive depth estimation performance has been achieved thanks to the availability of large-scale depth datasets~\cite{Chen_2020_CVPR, li2018megadepth, Xian2018DepthWeb,Zamir_2018_CVPR} and several technical
breakthroughs including innovative architecture designs \cite{fang2020towards, fu2018deep, huynh2020guiding, johnston2020self, li2017two, Poggi_2020_CVPR}, effective incorporation of geometric and semantics constraints \cite{chen2019towards, semanticconquerdepth, yang2020d3vo, yin2019enforcing, zhang2018joint, zhang2019pattern, zhu2020edge}, novel loss functions \cite{lossrebalancing, li2018megadepth, midas, wang2020cliffnet, xian2020sgr}, and supervision strategies \cite{depthattentionmodel, godard2017unsupervised, godard2019monodepth2, guo2018learning, jiang2018self, klingner2020self, li2020unsupervised, watson2019depthhints, wong2019bilateral}. 
In this work, rather than developing a new depth estimation method, we show that by merging estimations from different resolutions and patches, existing depth estimation models can be adapted to generate higher-quality results.


While impressive performance has been achieved across depth estimation benchmarks, most existing methods are trained to perform on relatively small input resolution, impeding their use in applications for which high-resolution depth maps are desirable \cite{niklaus20193d, shih20203d, wadhwa2018synthetic, wang2018deeplens}. 
Several works propose refinement methods for low-resolution depth estimates using guided upsampling alone~\cite{dziembowski2016depth, niklaus20193d} or in combination with residual training~\cite{zuo2020frequency}. 
Our approach instead focuses on generating the high-frequency details by changing the input of the network and merging multiple estimations. 

Our patch-based framework shares similarities with patch-based image editing, matting, and synthesis techniques where local results are generated from image patches and blended into global results \cite{darabi2012image, kalantari2014improving, yang2017high, yifan2019patch, yu2020high}. 
While related, existing patch-based editing techniques are not directly applicable to our scenario because of problem-specific challenges in monocular depth estimation. These challenges include varying range of depth values in patch estimates, strong dependency on context present in the image patch, and characteristic low-frequency artifacts that arise in high-resolution depth estimates.

\begin{figure*}\vspace{-0.1in}
\showimagew[\linewidth]{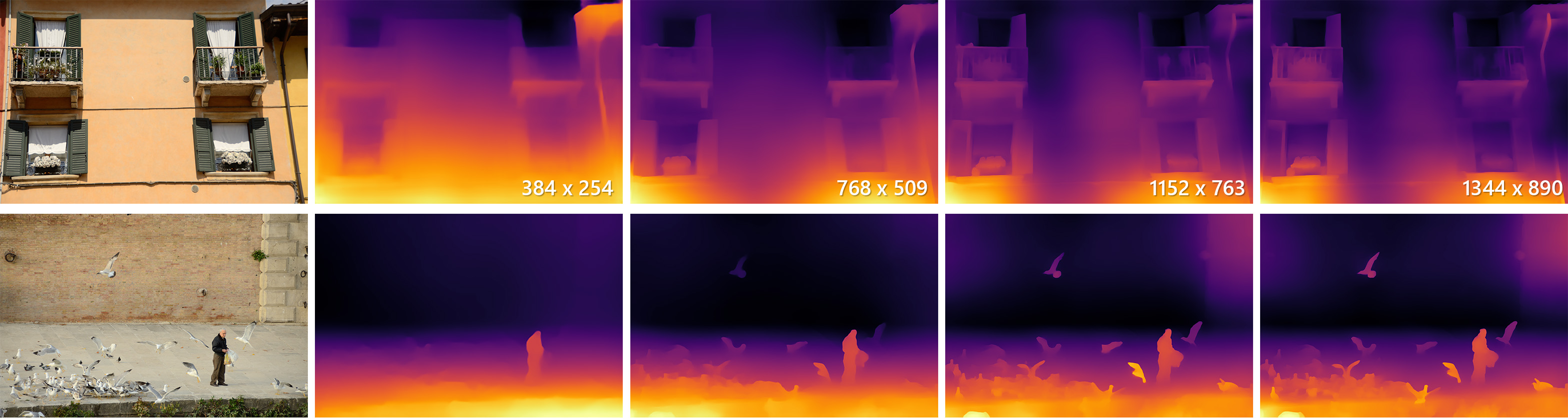}
\caption{
At small input resolutions, the network~\cite{midas} can estimate the overall structure of the scene successfully but often misses the details in the image, notice the missing birds in the bottom image. As the resolution gets higher, the performance around boundaries gets much better. However, the network starts losing the overall structure of the scene and generates low-frequency artifacts in the estimate. The resolution at which these artifacts start appearing depends on the distribution of contextual cues in the image. 
\negvspace
\negvspace
\negvspace
}
\label{fig:resolution}
\end{figure*}

\section{Observations on Model Behavior}
\label{sec:monocular}

\begin{figure}
\showimagew[\linewidth]{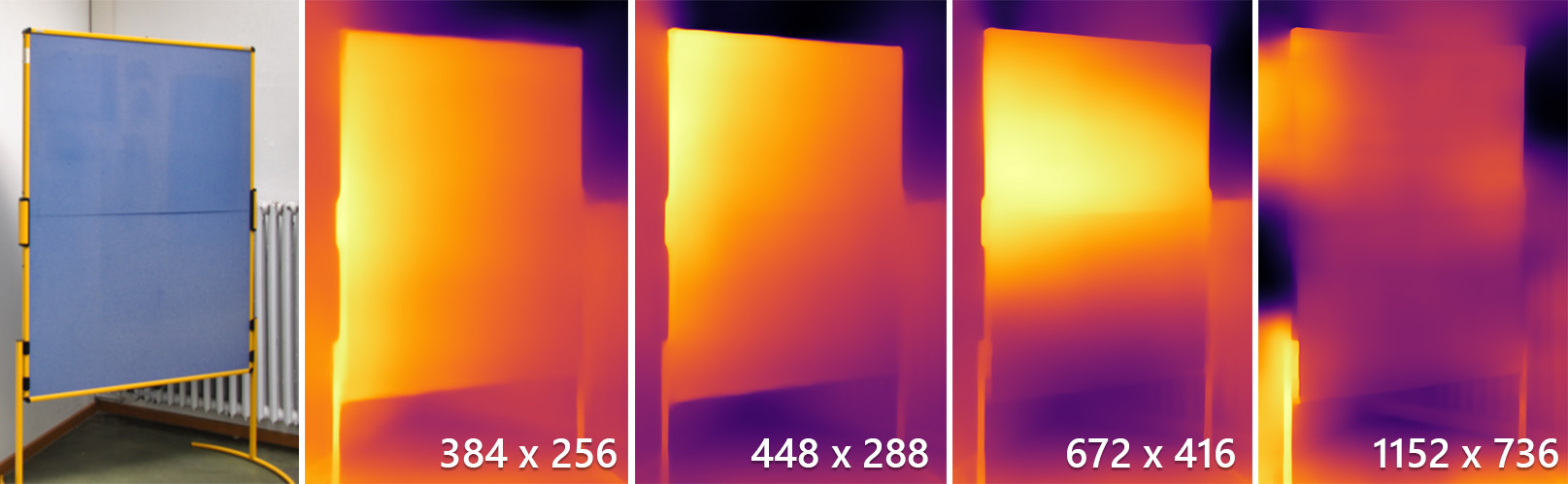}
\caption{
Since the model is fixed, changing the image resolution affects how much of the scene the receptive field can ``see''. As the resolution increases, depth cues get farther apart, starving the network of information, which progressively degrades the accuracy.
\negvspace
\negvspace
}
\label{fig:receptivefield}
\end{figure}

Monocular depth estimation, with the lack of geometric cues that multi-camera systems exploit, has to rely on high-level depth cues present in the image. 
In their analysis, Hu \etal~\cite{hu2019visualization} show that monocular depth estimation models indeed make use of monocular depth cues that the human visual system utilizes such as occlusions and perspective-related cues~\cite{SternbergVisualPerc} that we will refer to as contextual cues or more broadly as \emph{context}. 
In this section, we present our observations on how the context or more importantly how the \emph{context density} in the input affects the network performance. 
We present examples from MiDaS~\cite{midas} and show similar results from \cite{xian2020sgr} in the supplementary material.


Most depth estimation methods follow the common practice of training with a pre-defined and relatively low input resolution but the models themselves are fully convolutional, which in principle can handle arbitrary input sizes. 
When we feed an image into the same model with different resolutions, however, we see a specific trend in the result characteristics. 
Figure~\ref{fig:resolution} demonstrates that in smaller resolutions the estimations lack many high-frequency details while generating a consistent overall structure of the scene. 
As the input resolution gets higher, more details are generated in the result but we see inconsistencies in the scene structure characterized by gradual shifts in depth between image regions. 
We explain this duality through the limited capacity and the limited receptive field size of the network.

\begin{figure}
\showimagew[\linewidth]{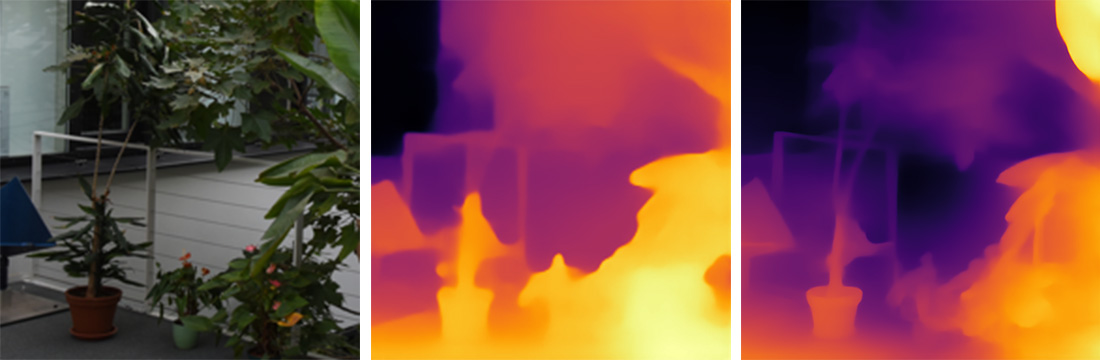}
\caption{
The original image with resolution $192\times192$ gains additional details in the depth estimate when fed to the network after upsampling to $500\times500$ (right) instead of its original resolution (middle).
\negvspace
\negvspace
\negvspace
}
\label{fig:upsampledEst}
\end{figure}

The receptive field size of a network depends mainly on the architecture as well as the training resolution. 
It can be defined as the region around a pixel that contributes to the output at that pixel~\cite{araujo2019computing,Le2017receptivefield}. 
As monocular depth estimation relies on contextual cues, when these cues in the image gets further apart than the receptive field, the network is not able to generate a coherent depth estimation around pixels that do not receive enough information. 
We demonstrate this behavior with a simple scene in Figure~\ref{fig:receptivefield}. 
MiDaS~\cite{midas} with its receptive field size of $384\times384$ starts to generate inconsistencies between image regions as the input gets larger and the contextual cues concentrated at the edges of the image get further apart than $384$ pixels. 
The inconsistent results for the flat wall in Figure~\ref{fig:resolution} (top) also support this observation. 

Convolutional neural networks have an inherently limited capacity that provides an upper bound to the amount of information they can store and generate~\cite{baldi2019capacity}. 
As the network can only \emph{see} as much as its receptive field size at once, the limit in capacity applies to what the network can generate inside its receptive field. 
We attribute the lack of high-frequency details in low-resolution estimates to this limit. 
When there are many contextual cues present in the input, the network is able to reason about the larger structures in the scene much better and is hence able to generate a consistent structure. 
However, this results in the network not being able to generate high-frequency details at the same time due to the limited amount of information that can be generated in a single forward pass. 
We show a simple experiment in Figure~\ref{fig:upsampledEst}. 
We use an original input image of $192\times192$ pixels and simply upsample it to generate higher resolution results. 
This way, the amount of high-frequency information remains the same in the input but we still see an increase in the high-resolution details in the result, demonstrating a limit in the network capacity.
We hence claim that the network gets \emph{overwhelmed} with the amount of contextual cues concentrated in a small image and is only able to generate an overall structure of the scene.

\section{Method Preliminaries}
\label{sec:preliminaries}

Following our observations in Section~\ref{sec:monocular}, our goal is to generate multiple depth estimations of a single image to be merged to achieve a result that has high-frequency details with a consistent overall structure. 
This requires (i) retrieving the distribution of contextual cues in the image that we will use to determine the inputs to the network, and (ii) a merging operation to transfer the high-frequency details from one estimate to another with structural consistency. 
Before going into the details of our pipeline in Sections~\ref{sec:double} and \ref{sec:patch}, we present our approach to these preliminaries. 

\negvspace
\paragraph{Estimating Contextual Cues}
Determining the contextual cues in the image is not a straightforward task. 
Hu et al.~\cite{hu2019visualization} focus on this problem by identifying the most relevant pixels for monocular depth estimation in an image. 
While they provide a comparative analysis of contextual cues used by different models during inference, we were not able to generate high-resolution estimates we need for cue distribution for MiDaS~\cite{midas} using their method. 
Instead, following their observation that image edges are reasonably correlated with the contextual cues, we use an approximate edge map of the image obtained by thresholding the RGB gradients as a proxy. 

\paragraph{Merging Monocular Depth Estimates}
In our problem formulation, we have two depth estimations that we would like to merge: (i) a low-resolution map obtained with a smaller-resolution input to the network and (ii) a higher-resolution depth map of the same image (Sec.~\ref{sec:double}) or patch (Sec.~\ref{sec:patch}) that has better accuracy around depth discontinuities but suffers from low-frequency artifacts. 
Our goal is to embed the high-frequency details of the second input into the first input which provides a consistent structure and a fixed range of depths for the full image. 

\begin{figure*}\vspace{-0.1in}
\showimagew[\linewidth]{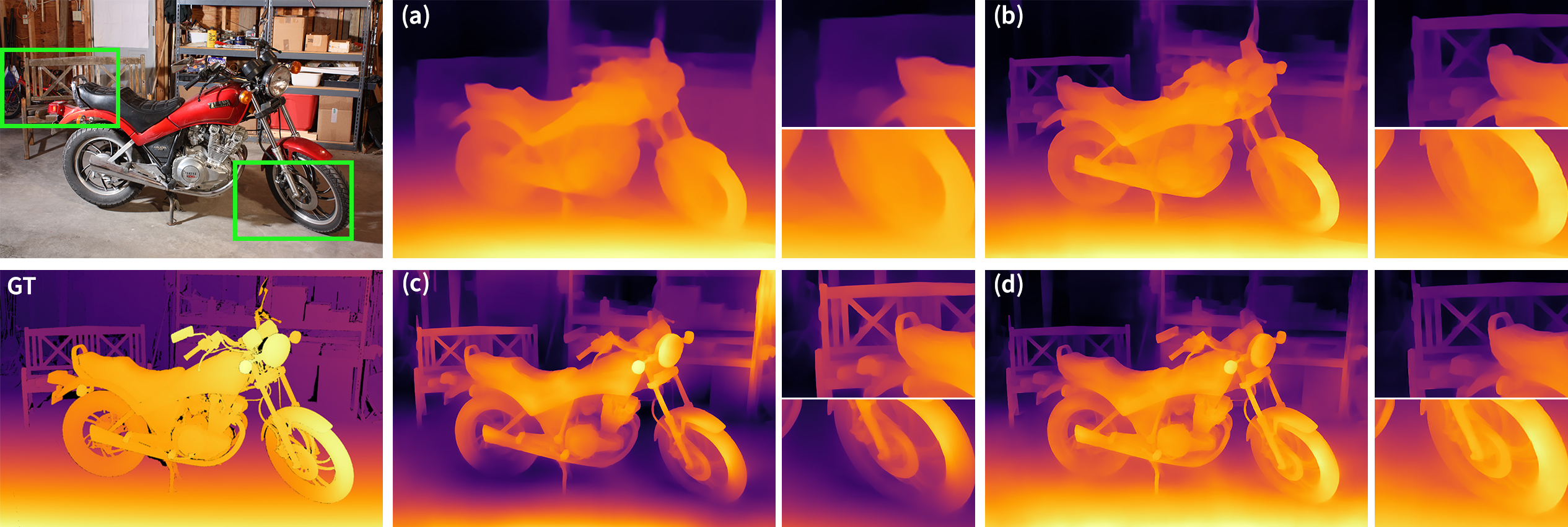}
\caption{
We show the depth estimates obtained at different resolutions, (a) at the training resolution of MiDaS~\cite{midas} at $384\times384$, (b) at the selected resolution with edges separated at most by 384 pixels, and (c) at a higher resolution that leaves 20\% of the pixels without nearby edges. Although the increasing resolution provides sharper results, beyond (c), the estimates become unstable in terms of the overall structure, visible through incorrect depth range for the bench in the background and unrealistic depth gradients around the tires. (d) Our merging network is able to fuse the fine-grain details in (c) into the consistent structure in (a) to get the best of two worlds.
}\vspace{-0.12in}
\label{fig:doubleest}
\end{figure*}

While this problem resembles gradient transfer methods such as Poisson blending~\cite{poisson}, due to the low-frequency artifacts in the high-resolution estimate, such low-level approaches do not perform well for our purposes. 
Instead, we utilize a standard network and adopt the Pix2Pix architecture~\cite{pix2pix} with a 10-layer U-net~\cite{unet} as the generator. 
Our selection of a 10-layer U-net instead of the default 6-layer aims to increase the training and inference resolution to $1024\times1024$, as we will use this merging network for a wide range of input resolutions. 
We train the network to transfer the fine-grained details from the high-resolution input to the low-resolution input. 
For this purpose, we generate input/output pairs by choosing patches from depth estimates of a selected set of images from Middlebury2014 \cite{Scharstein2014HighResolutionSD} and Ibims-1 \cite{Koch18}. 
While creating the low- and high-resolution inputs is not a problem, consistent and high-resolution ground truth cannot be generated natively. 
Note that we also can not directly make use of the original ground-truth data because we are training the network only for the low-level merging operation and the desired output depends on the range of depth values in the low-resolution estimate.
Instead, we empirically pick 672*672 pixels as input resolution to the network which maximizes the number of artifact-free estimations we can obtain over both datasets. To ensure that the ground truth and higher-resolution patch estimation have the same amount of fine-grained details, we apply a guided filter on the patch estimation using the ground truth estimation as guidance. These modified high-resolution patches serve as proxy ground truth for a seamlessly merged version of low- and high-resolution estimations. 
Figures~\ref{fig:doubleest} and \ref{fig:merging} demonstrate our merging operation.

\section{Double Estimation}
\label{sec:double}

We show the trade-off between a consistent scene structure and the high-frequency details in the estimates in Section~\ref{sec:monocular} and Figure~\ref{fig:resolution} with changing input resolution. 
We also show in Figures~\ref{fig:resolution} and \ref{fig:receptivefield} that the network starts to produce structurally inconsistent results when the contextual cues in the image are further apart than the receptive field size. 
The maximum resolution at which the network will be able to generate a consistent structure depends on the distribution of the contextual cues in the image. 
Using an edge map as the proxy for contextual cues, we can determine this maximum resolution by making sure that no pixel is further apart from contextual cues than half of the receptive field size. 
For this purpose, we apply binary dilation to the edge map with a receptive-field-sized kernel in different resolutions. 
Then, the resolution for which the dilated edge map stops to produce all-one results is the maximum resolution where every pixel will receive context information in a forward pass. 
We refer to this resolution that is adaptive to the image content as $\mathcal{R}_{0}$. 
We will refer to resolutions above $\mathcal{R}_{0}$ as $\mathcal{R}_x$ where $x$ represents the percentage of pixels that do not receive any contextual information at a given resolution.
Estimations with resolutions above $\mathcal{R}_{0}$ will lose structural consistency but they will have richer high-frequency content in the result.

Following these observations, we propose an algorithm that we call double estimation: to get the best of two worlds, we feed the image to the network in two different resolutions and merge the estimates to get a consistent result with high-frequency details. Our low-resolution estimation is set to the receptive field size of the network that will determine the overall structure in the image. 
Resolutions below the receptive field size do not improve the structure and in fact reduce the performance as the full capacity of the network is not utilized. 
We determined through experimental analysis in Section~\ref{sec:results:double} that our merging network can successfully merge the high-frequency details onto the low-resolution estimate's structure up to $\mathcal{R}_{20}$. 
The low-resolution artifacts in estimations beyond $\mathcal{R}_{20}$ start to damage the merged results. 
Note that $\mathcal{R}_{20}$ may be higher than the original resolution.

Figure~\ref{fig:doubleest} demonstrates that we can preserve the structure in the low-resolution estimation (a) while integrating the details in the high-resolution estimation (c) successfully into our result (d). 
Through merging, we can generate consistent results beyond $\mathcal{R}_{0}$ (b), which is the limit set by the receptive field size of the network, at the cost of a second forward-pass through the base network.

\section{Patch Estimates for Local Boosting}
\label{sec:patch}

We determine the estimation resolution for the whole image based on the number of pixels that do not have any contextual cues nearby. These regions with the lowest contextual cue density are dictating the maximum resolution we can use for an image. 
Regions with higher contextual cue density, however, would still benefit from higher-resolution estimations to generate more high-frequency details. 
We present a patch-selection method to generate depth estimates at different resolutions for different regions in the image that are merged together for a consistent full result. 

Ideally, the patch selection process should be guided with high-level information that determines the local resolution optimum for estimation. 
This requires a data-driven approach that can evaluate the high-resolution performance of the network and an accurate high-resolution estimation of the contextual cues. 
However, the resolution of the currently available datasets are not enough to train such a system. 
As a result, we present a simple patch selection method where we make cautious design decisions to arrive at a reliable high-resolution depth estimation pipeline without requiring an additional dataset or training. 
\paragraph{Base estimate}
We first generate a base estimate using the double estimation in Section~\ref{sec:double} for the whole image. 
The resolution of this base estimate is fixed as $\mathcal{R}_{20}$ for most images. Only for a subset of images we increase this resolution as detailed at the end of this section. 

\paragraph{Patch selection}
We start the patch selection process by tiling the image at the base resolution with a tile size equal to the receptive field size and a 1/3 overlap. 
Each of these tiles serves as a candidate patch. 
We ensure each patch receives enough context to generate meaningful depth estimates by comparing the density of the edges in the patch to the density of the edges in the whole image. 
If a tile has less edge density than the image, it is discarded. 
If a tile has a higher edge density, the size of the patch is increased until the edge density matches the original image. 
This makes sure that each patch estimate has a stable structure.

\begin{figure}
\showimagew[\linewidth]{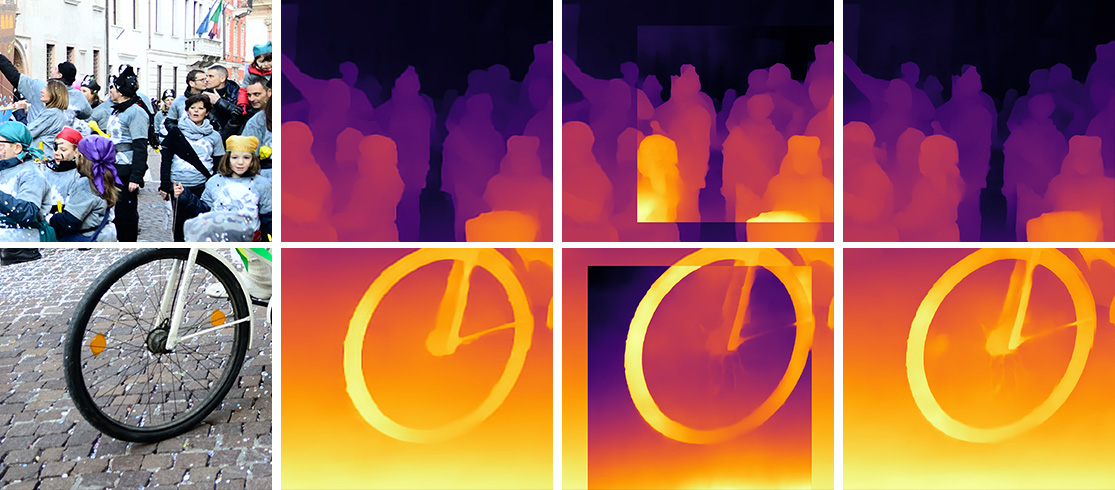}
\caption{
Input patches are shown in our base estimate, patch-estimate pasted onto the base estimate, and our result after merging. The image is picked from~\cite{dang2015raise}.
\negvspace
}\vspace{-0.12in}
\label{fig:merging}
\end{figure}

\paragraph{Patch estimates}
We generate depth estimates for patches using another double estimation scheme. 
Since the patches are selected with respect to the edge density, we do not adjust the estimation resolution further. Instead, we fix the high-resolution estimation size to double the receptive field size. 
The generated patch-estimates are then merged onto this base estimate one by one to generate a more detailed depth map as shown in Figure~\ref{fig:pipeline}. 
Note that the range of depth values in the patch estimates differs from the base estimate since monocular depth estimation networks do not provide a metric depth. Rather their results represent the ordinal depth relationship between image regions. 
Our merging network is designed to handle such challenges and can successfully merge the high-frequency details in the patch estimate onto the base estimate as Figure~\ref{fig:merging} shows.

\paragraph{Base resolution adjustment} 
We observe that when the edge density in the image varies a lot, especially when a large portion of the image lacks any edges, our patch selection process ends up selecting too small patches due to the small $\mathcal{R}_{20}$. 
We solve this issue for such cases by upsampling the base estimate to a higher resolution before patch selection. 
For this, we first determine a maximum base size $\mathcal{R}_{max} = 3000\times3000$ from the $\mathcal{R}_{20}$ value of a 36-megapixel outdoors image with a lot of high-frequency content. 
Then we define a simple multiplier for the base estimate size as 
$\max\left( 1, {\mathcal{R}_{max}}/{(4\mathcal{K}\mathcal{R}_{20})} \right)$ 
%
%
where $\mathcal{K}$ is the percentage of pixels in the image that are close to edges determined by dilating the edge map with a kernel of quarter of the size of the receptive field. 
The $\max$ operation makes sure that we never decrease the base size. 
This multiplier makes sure that we can select small high-density areas within an overall low-density image with small patches when we define the minimum patch size as the receptive field size in the base resolution.

\section{Results and Discussion}
\label{sec:results}
We evaluate our method on two different datasets, Middleburry 2014~\cite{Scharstein2014HighResolutionSD} for which high-resolution inputs and ground-truth depth maps are available, and IBMS-1~\cite{Koch18}. We evaluate using a set of standard depth evaluation metrics as suggested in recent work~\cite{midas,xian2020sgr}, including root mean squared error in disparity space (RMSE), percentage of pixels with $\delta = \max(\frac{z_i}{z_i^*},\frac{z_i^*}{z_i}) > 1.25$ ($\delta_{1.25}$), and ordinal error (ORD) from~\cite{xian2020sgr} in depth space. 
Additionally, we propose a variation of ordinal relation error~\cite{xian2020sgr} that we call depth discontinuity disagreement ratio (D$^3$R) to measure the quality of high frequencies in depth estimates. Instead of using random points as in~\cite{xian2020sgr} for ordinal comparison, we use the centers of superpixels~\cite{slic} computed using the ground truth depth and compare neighboring superpixel centroids across depth discontinuities. This metric hence focuses on boundary accuracy. 
We provide a more detailed description of our metric in the supplementary material.

Our merging network is light-weight and the time it takes to do a forward pass is magnitudes smaller than the monocular depth estimation networks. 
The running time of our method mainly depends on how many times we use the base network in our pipeline. 
The resolution at which the base estimation is computed, $\mathcal{R}_{20}$, and the number of patches we merge onto the base estimate is adaptive to the image content. 
Our method ended up selecting 74.82 patches per image on average with an average  $\mathcal{R}_{20}=2145\times1501$ for the Middleburry 2014~\cite{Scharstein2014HighResolutionSD} dataset and 12.17 patches per image with an average  $\mathcal{R}_{20}= 1443\times1082$ for IBMS-1~\cite{Koch18}. 
The difference between these numbers comes from the different scene structures present in the two datasets. 
Also note that the original image resolution of IBMS-1~\cite{Koch18} is $640\times480$. 
As we demonstrate in Section~\ref{sec:monocular}, upscaling low-resolution images does help in generating more high-frequency details. 
Hence, our estimation resolution depends mainly on the image content and not on the original input resolution.

\begin{table*}[t]
\caption{
    Quantitative evaluation of our method using two base networks on two different datasets. Lower is better.
}
\negvspace
\resizebox{\linewidth}{!}
{
{
\setlength{\extrarowheight}{4pt}
\setlength\arrayrulewidth{1pt}
\rowcolors{2}{gray!25}{white}

\begin{tabular}{!{\color{gray!25}\vrule} l!{\color{gray!25}\vrule} !{\color{gray!25}\vrule} cccc!{\color{gray!25}\vrule} cccc!{\color{gray!25}\vrule} !{\color{gray!25}\vrule} cccc!{\color{gray!25}\vrule} cccc!{\color{gray!25}\vrule} }
    \arrayrulecolor{gray!25}\hline
     & \multicolumn{8}{c!{\color{gray!25}\vrule} !{\color{gray!25}\vrule} }{Middleburry2014\cite{Scharstein2014HighResolutionSD}} & \multicolumn{8}{c!{\color{gray!25}\vrule} }{Ibims-1~\cite{Koch18}}  \\ 
    \arrayrulecolor{gray!25}\hline
     & \multicolumn{4}{c!{\color{gray!25}\vrule} }{MiDaS~\cite{midas}} & \multicolumn{4}{c!{\color{gray!25}\vrule} !{\color{gray!25}\vrule} }{SGR~\cite{xian2020sgr}} & \multicolumn{4}{c!{\color{gray!25}\vrule} }{MiDaS~\cite{midas}} & \multicolumn{4}{c!{\color{gray!25}\vrule} }{SGR~\cite{xian2020sgr}}  \\ 
    \arrayrulecolor{gray!25}\hline 
     & ORD & D$^3$R & RMSE & $\delta_{1.25}$ 
                & ORD & D$^3$R & RMSE & $\delta_{1.25}$ 
                    & ORD & D$^3$R & RMSE & $\delta_{1.25}$  
                        & ORD & D$^3$R & RMSE & $\delta_{1.25}$ \\
                        
    Original Method                         & 0.3840 & 0.3343 & 0.1708 & 0.7649                     
                                                & 0.4087 & 0.3889 & 0.2123 & 0.7989 
                                                    & 0.4002 & 0.3698 & \topscore{0.1596} & \topscore{0.6345}
                                                        & 0.5555 & 0.4736 & 0.1956 & 0.7513\\ 
    Refine-Bilateral                    & 0.3806 & 0.3366 & 0.1707 & 0.7627
                                                & 0.4078 & 0.3904 & 0.2122 & 0.7990
                                                    & 0.3982 & 0.3768 & 0.1596 & 0.6350 
                                                        & 0.5551 & 0.4750 & 0.1956 & 0.7501\\    
    Refine-with~\cite{niklaus20193d}    & 0.3826 & 0.3377 & 0.1704 & 0.7622 
                                                & 0.4081 & 0.3880 & 0.2115 & 0.7993 
                                                    & 0.4006 & 0.3761 & 0.1600 & 0.6351
                                                        & 0.5488 & 0.4780 & \topscore{0.1953} & 0.7482\\ 
    Single-est ($\mathcal{R}_{0}$)              & 0.3554 & 0.2504 & \topscore{0.1481} & \topscore{0.7161}
                                                & 0.4312 & 0.3131 & 0.1999 & \topscore{0.7841} 
                                                    & 0.4504 & 0.3269 & 0.1687 & 0.6633
                                                        & 0.6343 & 0.4901 & 0.2146 & 0.7856\\  
    Double-est ($\mathcal{R}_{20}$)              & 0.3496 & 0.1709 & 0.1563 & 0.7364                    
                                                & 0.3944 & 0.2540 & 0.1983 & 0.7931 
                                                    & 0.4112 & 0.3272 & 0.1597 & 0.6386
                                                        & 0.5591 & 0.4829 & 0.1967 & 0.7473\\
    OURS                                    & \topscore{0.3467} & \topscore{0.1578} & 0.1557 & 0.7406  
                                                & \topscore{0.3879} & \topscore{0.2324} & \topscore{0.1973} & 0.7891
                                                    & \topscore{0.3938} & \topscore{0.3222} & 0.1598 & 0.6390
                                                         & \topscore{0.5538} & \topscore{0.4671} & 0.1965 & \topscore{0.7460}\\
    \arrayrulecolor{gray!25}\hline 
\end{tabular}
}
}
\negvspace
\label{tab:depth_estimation_performance1}
\end{table*}

\begin{figure*}
{
\footnotesize
\begin{tabular}
{   K{0.175\linewidth}
    K{0.175\linewidth}
    K{0.175\linewidth}
    K{0.175\linewidth}
    K{0.175\linewidth}
}
Input&
MiDaS~\cite{midas}&
Ours using MiDaS&
SGR~\cite{xian2020sgr}&
Ours using SGR
\end{tabular}
}
\showimagew[\linewidth]{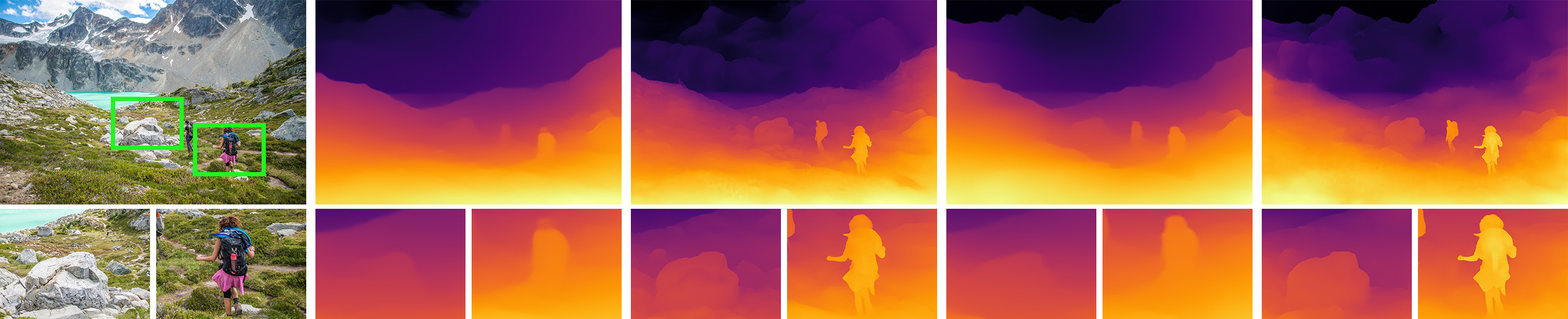}
\\
\showimagew[\linewidth]{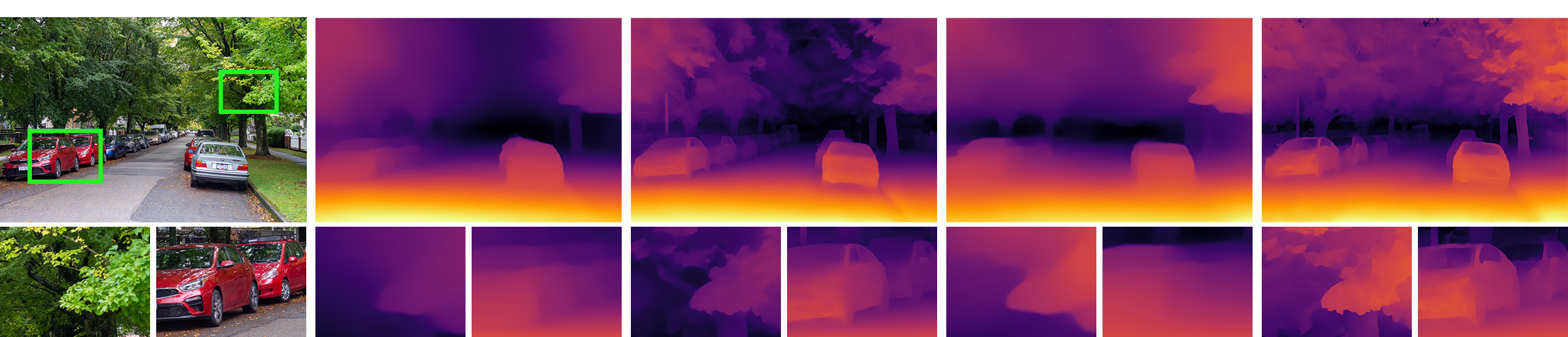}
\\
\showimagew[\linewidth]{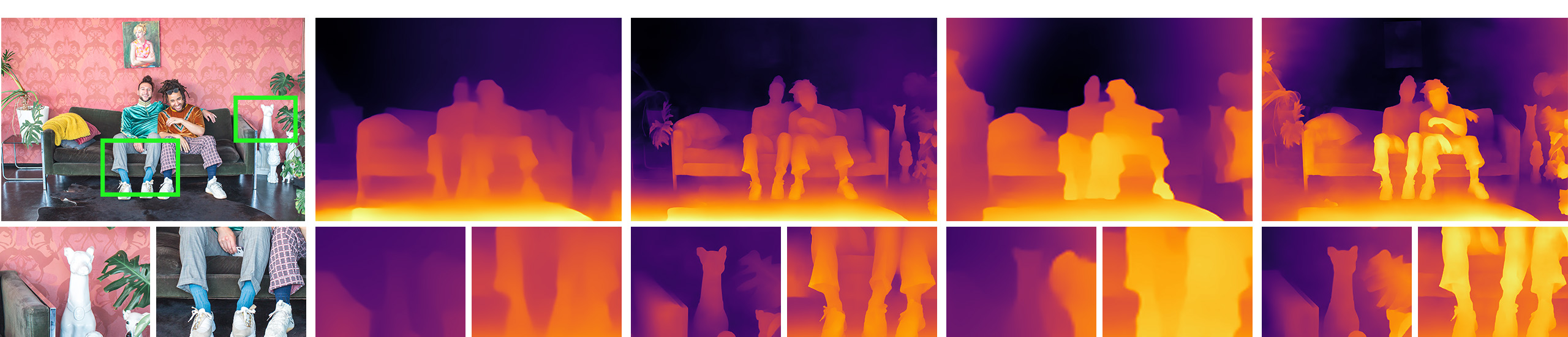}
\caption{
Additional results using MiDaS~\cite{midas} and the Structure-Guided Ranking Loss method~\cite{xian2020sgr} compared to the original methods run at their default size.
\negvspace
\negvspace
}
\label{fig:MainComp}
\end{figure*}

\begin{figure*}
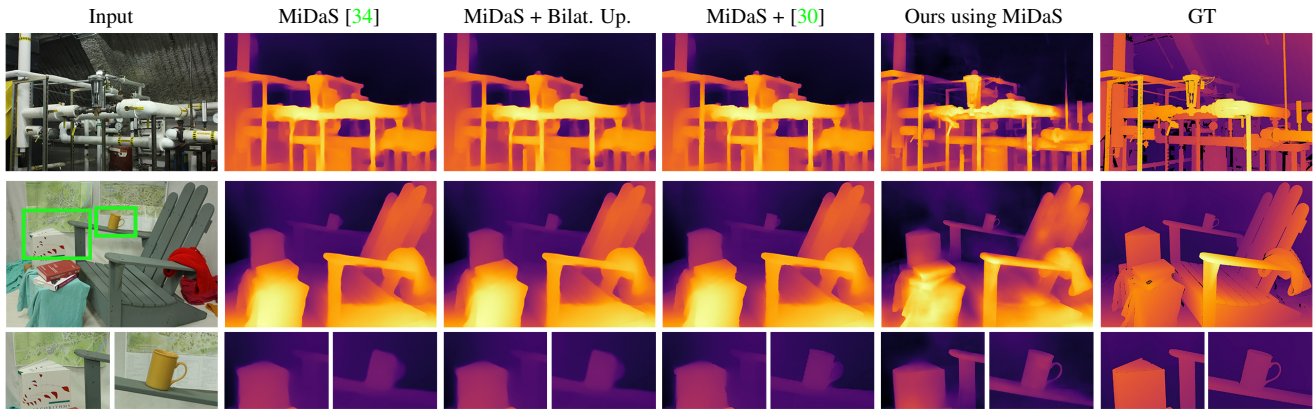
\vspace{-0.2in}
{
\footnotesize
\begin{tabular}
{   K{0.142\linewidth}
    K{0.142\linewidth}
    K{0.142\linewidth}
    K{0.142\linewidth}
    K{0.142\linewidth}
    K{0.142\linewidth}
}
Input&
MiDaS~\cite{midas}&
MiDaS + Bilat. Up.&
MiDaS + \cite{niklaus20193d}&
Ours using MiDaS &
GT
\end{tabular}
}
\showimagew[\linewidth]{images/refinementComp2}
\caption{
We compare out method to bilateral upsampling and the refinement method proposed by Niklaus et al.~\cite{niklaus20193d} as applied to MiDaS~\cite{midas} output. Refinement methods fail to add any details that do not exist in the original estimation. With our patch-based merging framework, we are able to generate sharp details in the image.
}
\label{fig:refinementComp}
\end{figure*}

\subsection{Boosting Monocular Depth Estimation Models}
We evaluate how much our method can improve upon pre-trained monocular depth estimation models using MiDaS~\cite{midas} and SGR~\cite{xian2020sgr} as well as the depth refinement method by Niklaus et al.~\cite{niklaus20193d} and a baseline where we refine the original method's results using a bilateral filter after bilinear upsampling. 
The quantitative results in Table~\ref{tab:depth_estimation_performance1} show that for the majority of the metrics, our full pipeline improves the numerical performance considerably and our double-estimation method already provides a good improvement at a small computational overhead. 
Our content-adaptive boosting framework consistently improves the depth estimation accuracy over the baselines on both datasets in terms of ORD and D$^3$R metrics, indicating accurate depth ordering and better-preserved boundaries. Our method also performs comparably in terms of RMSE and $\delta_{1.25}$.
We also observe that simply adjusting the input resolution adaptively to $\mathcal{R}_{0}$ meaningfully increases the performance. 

The performance improvement provided by our method is much more significant in qualitative comparisons shown in Figure~\ref{fig:MainComp}. 
We can drastically increase the number of high-frequency details and the boundary localization when compared to the original networks. 

We do not see a large improvement when depth refinement methods are used in Table~\ref{tab:depth_estimation_performance1} and also in the qualitative examples in Figure~\ref{fig:refinementComp}. 
This difference comes from the fact that we utilize the network multiple times to generate richer information while the refinement methods are limited by the details available in the base estimation results. 
Qualitative examples show that the refinement methods are not able to generate additional details that were missed in the base estimate such as small objects or sharp depth discontinuities. 

\begin{table}[t]
\caption{
    Whole image estimation performance of MiDaS~\cite{midas} with changing resolution and double estimation on the Middlebury dataset~\cite{Scharstein2014HighResolutionSD}. Lower is better.
}
\resizebox{\linewidth}{!}
{
{
\setlength{\extrarowheight}{8pt}
\setlength\arrayrulewidth{1pt}
\rowcolors{2}{gray!25}{}
\begin{tabular}
{!{\color{gray!25}\vrule}l!{\color{gray!25}\vrule}!{\color{gray!25}\vrule}cccc!{\color{gray!25}\vrule}!{\color{gray!25}\vrule}ccc!{\color{gray!25}\vrule}!{\color{gray!25}\vrule}cccc!{\color{gray!25}\vrule}}
    \arrayrulecolor{gray!25}\hline 
     & \multicolumn{7}{c!{\color{gray!25}\vrule}!{\color{gray!25}\vrule}}{Single estimation} & \multicolumn{4}{c!{\color{gray!25}\vrule}}{Double estimation} \\
     \arrayrulecolor{gray!25}\hline
     & \multicolumn{4}{c!{\color{gray!25}\vrule}!{\color{gray!25}\vrule}}{Fixed size (pixels)} & \multicolumn{7}{c!{\color{gray!25}\vrule}}{Context-adaptive} \\
     \arrayrulecolor{gray!25}\hline 
     & 384 & 768 & 1152 & 1536 & $\mathcal{R}_{0}$ & $\mathcal{R}_{10}$ & $\mathcal{R}_{20}$ & $\mathcal{R}_{0}$ & $\mathcal{R}_{10}$ & $\mathcal{R}_{20}$ & $\mathcal{R}_{30}$ \\
     ORD  
        & 0.384 & 0.371 & 0.426 & 0.478 & 0.355 & 0.457 & 0.505 & 0.361 & \topscore{0.349} & \topscore{0.349} & 0.352 \\
     D$^3$R 
        & 0.334 & 0.217 & 0.187 & 0.189 & 0.250 & 0.197 & 0.199 & 0.258 & 0.183 & \topscore{0.170} & 0.171 \\
     RMSE
        & 0.170 & 0.152 & 0.165 & 0.186 & \topscore{0.148} & 0.183 & 0.198 & 0.164 & 0.157 & 0.156 & 0.156 \\
     {$\delta_{1.25}$} 
        & 0.764 & 0.745 & 0.740 & 0.793 & \topscore{0.716} & 0.788 & 0.803 & 0.749 & 0.730 & 0.736 & 0.745 \\
    \arrayrulecolor{gray!25}\hline 
\end{tabular}

}
}
\label{tab:doubleest}
\end{table}

\subsection{Double Estimation and $\mathcal{R}_{x}$}
\label{sec:results:double}
We chose $\mathcal{R}_{20}$ as the high-resolution estimation in our double-estimation framework. 
This number is chosen based on the quantitative results in Table~\ref{tab:doubleest}, where we show that using a higher resolution $\mathcal{R}_{30}$ results in a decrease in performance. 
This is due to the high-resolution results having heavy artifacts as the number of pixels in the image without contextual information increases. 
Table~\ref{tab:doubleest} also demonstrates that our double estimation framework outperforms fixed input resolutions which are the common practice, as well as estimations at $\mathcal{R}_{0}$ which represents the maximum resolution an image can be fed to the networks without creating structural inconsistencies. 

\subsection{Limitations}
\label{sec:results:limits}
Since our method is built upon monocular depth estimation, it suffers from its inherent limitations and therefore generates relative, ordinal depth estimates but not absolute depth values. 
We also observed that the performance of the base models degrade with noise and our method is not able to provide meaningful improvement for noisy images. We address this in an analysis on the NYUv2~\cite{Silberman:ECCV12} dataset in the supplementary material.
The high-frequency estimates suffer from low-magnitude white noise which is not always filtered out by our merging network and may result in flat surfaces appearing noisy in our results.

We utilize RGB edges as a proxy for monocular depth cues and make some ad-hoc choices in our patch selection process. 
While we are able to significantly boost base models with our current formulation, we believe research on contextual cues and the patch selection process will be beneficial to reach the full potential of pre-trained monocular depth estimation networks. 

\begin{figure}
\showimagew[\linewidth]{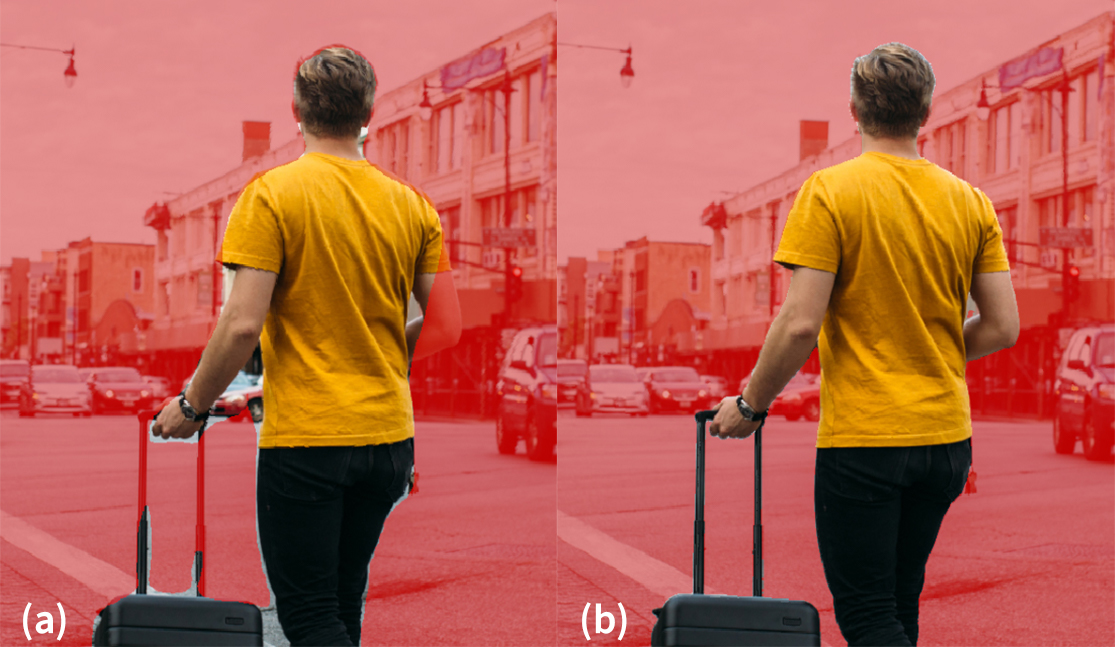}
\caption{
Our boundary accuracy is better visible in this example where we apply a threshold to the estimated depth values of MiDaS~\cite{midas} (a) and ours (b).
\negvspace
}
\label{fig:edgeaccuracy}
\negvspace
\end{figure}

\section{Conclusion}
\label{sec:conclusion}
We have demonstrated an algorithm to infer a high-resolution depth map from a single image using pre-trained models. While previous work is limited to sub-megapixel resolutions, our technique can process the multi-megapixel images captured by modern cameras.
High-quality high-resolution monocular depth estimation enables many application scenarios such as image segmentation. 
We show a simple segmentation by thresholding the depth values in Figure~\ref{fig:edgeaccuracy} which also demonstrates our boundary localization.
Our work is based on a careful characterization of the abilities of existing depth-estimation networks and the factors that influence them. We hope that our approach will stimulate more work on high-resolution depth estimation and pave the way for compelling applications.

{\small
\bibliographystyle{ieee_fullname}
\bibliography{depth}
}

\end{document}